\documentclass[10pt,twocolumn,letterpaper]{article}

\usepackage{cvpr}

\usepackage{graphicx}
\usepackage{amsmath}
\usepackage{amssymb}
\usepackage{booktabs}
\usepackage{xcolor}
\usepackage{soul}

\usepackage[pagebackref,breaklinks,colorlinks,citecolor=citecolor]{hyperref}
\definecolor{citecolor}{RGB}{34,139,34}

\newcommand{\app}{\raise.17ex\hbox{$\scriptstyle\sim$}}

\newcommand{\ourparagraph}[1]{\vspace{1.5mm}\noindent\textbf{#1}}

\newcommand{\tablestyle}[2]{\setlength{\tabcolsep}{#1}\renewcommand{\arraystretch}{#2}\centering\footnotesize}
\newlength\savewidth\newcommand\shline{\noalign{\global\savewidth\arrayrulewidth
  \global\arrayrulewidth 1pt}\hline\noalign{\global\arrayrulewidth\savewidth}}

\newcommand{\customfootnotetext}[2]{{
  \renewcommand{\thefootnote}{#1}
  \footnotetext[0]{#2}}}

\def\cvprPaperID{4433}
\def\confName{CVPR}
\def\confYear{2023}

\begin{document}
\title{Learning to Imitate Object Interactions from Internet Videos}
\author{%
 Austin Patel$^{*}$ \quad Andrew Wang$^{*}$ \quad Ilija Radosavovic \quad Jitendra Malik\\[2mm]
 University of California, Berkeley}
%\maketitle
\twocolumn[{%
\renewcommand\twocolumn[1][]{#1}%
\maketitle
\begin{center}
\vspace{-6mm}
\centerline{\includegraphics[width=1.0\linewidth]{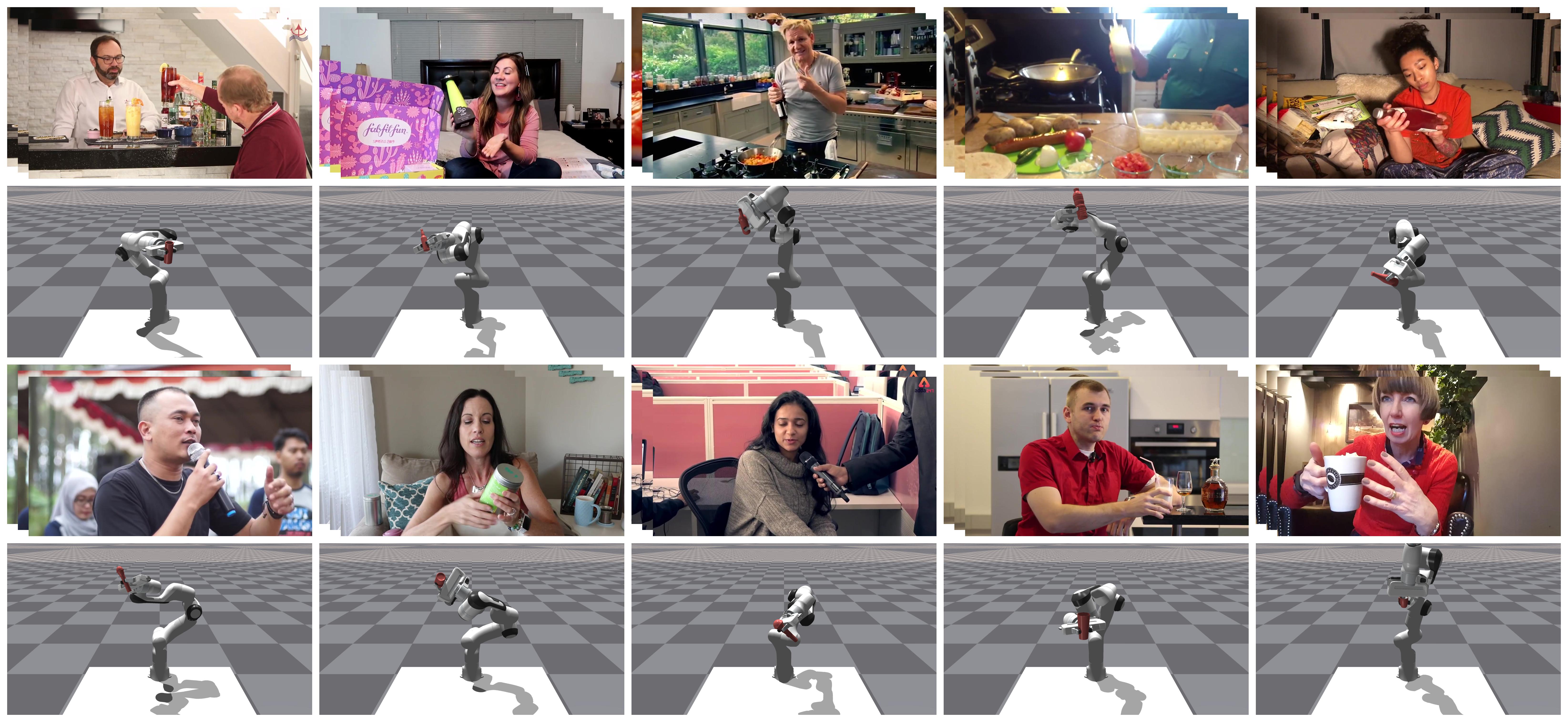}}
\vspace{-1mm}
\captionof{figure}{\small \textbf{Imitating object interactions from internet videos.} We study the problem of imitating object interactions from Internet videos. We present an approach that is able to imitate a range of different object interactions. Please see our~\href{https://austinapatel.github.io/imitate-video}{project page} for video results.}\label{fig:teaser}
\vspace{-2mm}
\end{center}%
}]

\customfootnotetext{*}{Equal contribution. Videos are available on our \href{https://austinapatel.github.io/imitate-video}{project page}.}

%%%%%%%%%%%%%%%%%%%%%%%%%%%%%%%%%%%%%%%%%%%%%%%%%%%%%%%%%%%%%%%%%%%%%%%%%%%%%%%%%%%%%%%%%%%%%%%%%%%
\begin{abstract}
\vspace{-2mm}
We study the problem of imitating object interactions from Internet videos. This requires understanding the hand-object interactions in 4D, spatially in 3D and over time, which is challenging due to mutual hand-object occlusions. In this paper we make two main contributions: (1) a novel reconstruction technique RHOV (Reconstructing Hands and Objects from Videos), which reconstructs 4D trajectories of both the hand and the object using 2D image cues and temporal smoothness constraints; (2) a system for imitating object interactions in a physics simulator with reinforcement learning. We apply our reconstruction technique to 100 challenging Internet videos. We further show that we can successfully imitate a range of different object interactions in a physics simulator. Our object-centric approach is not limited to human-like end-effectors and can learn to imitate object interactions using different embodiments, like a robotic arm with a parallel jaw gripper.
%\vspace{-1mm}
\end{abstract}

%%%%%%%%%%%%%%%%%%%%%%%%%%%%%%%%%%%%%%%%%%%%%%%%%%%%%%%%%%%%%%%%%%%%%%%%%%%%%%%%%%%%%%%%%%%%%%%%%%%
\section{Introduction}
\vspace{0mm}

Learning from visual demonstrations has become a popular problem in robotics~\cite{sermanet2018time,zhang2018deep,Qin2022dexmv,Bahl2022human,Radosavovic2022,Arunachalam2022holo}. These approaches have shown that we can leverage modern computer vision techniques to understand human actions from video streams and extract sufficient information for constructing demonstrations for robotics. However, in all of these cases the human is performing actions specifically for the robot to imitate. Most commonly, in a lab environment with instrumentation, such as depth cameras.

Wouldn't it be nice if we could instead imitate videos of normal human activity? We now have access to large amounts of video data from the Internet and computer vision datasets~\cite{Damen2018,100doh,Grauman2021} showing everyday human actions. This data is abundant and could potentially serve as a scalable source of realistic data for robot learning.

%##################################################################################################
\begin{figure*}[t]\centering\vspace{0mm}
\includegraphics[width=1.0\linewidth]{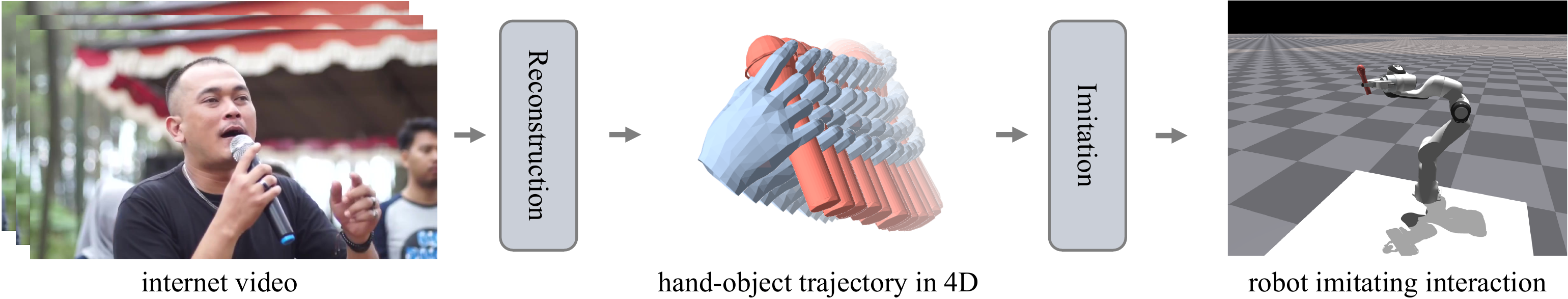}
\caption{\textbf{Approach.} We present an approach for imitating object interactions from Internet videos. We first reconstruct hand-object trajectories in 4D (see~\S\ref{sec:rhov}). We then use the recovered trajectories to imitate the object motion with a robot in a physics simulator (see~\S\ref{sec:rl}).}
\label{fig:overview}\vspace{-3mm}
\end{figure*}
%##################################################################################################

In this paper, we explore learning to imitate object interactions from Internet videos. This setting brings a number of challenges. First, we must be able to deal with Internet videos that contain activity in unconstrained everyday settings. Second, we require a computer vision technique that is able to recover hand-object interactions in 4D, spatially in 3D plus time, which is hard due to mutual hand-object occlusions. Third, reconstructing the perceived motion is not enough; we must also show that the motion can be imitated by a robot in a physics simulator.

To tackle this setting, we make two main contributions: (1) we develop a new technique for reconstructing hand-object trajectories from Internet videos, called RHOV for Reconstructing Hands and Objects from Videos; (2) we use this technique in conjunction with reinforcement learning to develop a system for imitating the reconstructed object trajectory in a physics simulator with a robot.

Specifically, RHOV is a new optimization-based technique for reconstructing hand-object trajectories from Internet videos in 4D. The basic idea is to leverage 2D image cues and temporal smoothness constraints. RHOV consists of three steps: hand reconstruction using hand keypoints and masks; object reconstruction via differentiable rendering with 2D mask and depth constraints; and temporal optimization with smoothness constraints in 3D space as well as in 2D image space via optical flow.

Given a 4D hand-object trajectory recovered by RHOV, we imitate the object interactions in a physics simulator. In particular, the recovered trajectory serves as a geometric goal specification that we use to compute a dense reward function. We then train a neural network policy to control a robot arm to pick up the object and imitate the object trajectory from the video. Our object-centric approach is not limited to human-like end-effectors and enables us to use a standard robotic arm with a parallel jaw gripper. The overall process can be seen as a way of improving kinematic 4D reconstructions by enforcing physics constraints in a simulator to obtain dynamic 4D reconstructions.

We apply our reconstruction approach to 100 YouTube videos from the 100 Days of Hands dataset~\cite{100doh}, that have now been ''4Dfied'' with help of our technique. We further show that we are able to imitate a range of different object interactions using a 7 DoF Franka robotic arm with a parallel jaw gripper (Figure~\ref{fig:teaser}). All of our code and recovered trajectories will be made publicly available.

%%%%%%%%%%%%%%%%%%%%%%%%%%%%%%%%%%%%%%%%%%%%%%%%%%%%%%%%%%%%%%%%%%%%%%%%%%%%%%%%%%%%%%%%%%%%%%%%%%%
\section{Related Work}

\ourparagraph{Hand-object reconstruction.} There is a large body of literature on reconstructing hands~\cite{sharp2015accurate, sridhar2013interactive, tagliasacchi2015robust, tzionas2016capturing, ye2016spatial, yuan2018depth, moon2018v2v,romero2010hands, mueller2018ganerated, cai2018weakly, yang2019disentangling, zimmermann2017learning, Kulon_2020_CVPR,frankmocap}, objects~\cite{cao2016real, factored3dTulsiani17, kundu20183d, Gkioxari2019, kuo2020mask2cad, lim2013parsing, Xiang2018, zhang2020phosa, Sun2018}, and both hands and objects~\cite{oikonomidis2011full,tzionas2016capturing,hasson19_obman,hasson20_handobjectconsist,rhoi,homan,ye2022s}. Majority of these works primarily focus on videos in lab settings or images in the wild. We build on these efforts but focus on reconstructing hand-object trajectories from Internet videos, which has received modest attention. Specifically, we extend the RHO technique from~\cite{rhoi} to videos by introducing a temporal optimization stage. Our approach is also related to~\cite{homan} who employ an optimization-based approach in a similar spirit but use a different set of loss functions (most notably we use depth and optical flow losses).

\ourparagraph{Learning to imitate with reinforcement learning.} Using reinforcement learning (RL) has been very effective for imitating motion capture trajectories~\cite{peng2018deepmimic}. RL-based imitation has also proven effective for noisier sources of trajectories, such as head-worn cameras~\cite{yuan20183d} and VR headsets~\cite{winkler2022questsim}. Most similar to ours is the work of~\cite{Peng2018sfv}, who imitate videos of humans, and~\cite{christen2022d}, who imitate static hand-object poses. Likewise, we use an RL-based imitation technique but focus on imitating object interactions from Internet videos.

\ourparagraph{Retargeting and teleoperation.} A number of works have explored using computer vision-based hand pose estimation for controlling or teleoperatic a robotic system~\cite{Handa2020,Arunachalam2022,Qin2022,sivakumar2022robotic,Arunachalam2022holo}. These approaches typically do not include the object and focus on human-like end-effectors. Instead, we take an object-centric perspective and focus on objects, which enables us to imitate object interactions with robots of different morphologies, like a robotic arm with a gripper.

\ourparagraph{Learning from demonstrations.} Learning from visual demonstrations is a popular topic in robotics~\cite{sermanet2018time,zhang2018deep,yu2018one,mandlekar2018roboturk,petrik2021learning,Qin2022dexmv,Bahl2022human,Radosavovic2022,Arunachalam2022holo}. These techniques are sometimes similar to the teleoperation ones in the types of  vision methods they use, but focus on using demonstrations for learning tasks with a robot. They too typically record human videos in the lab settings. Instead, we focus on imitating natural object interactions from Internet videos. In principle, our trajectories could be used as demonstrations as well~\cite{Radosavovic2021}.

%##################################################################################################
\begin{figure*}[t]\centering\vspace{0mm}
\includegraphics[width=0.80\linewidth]{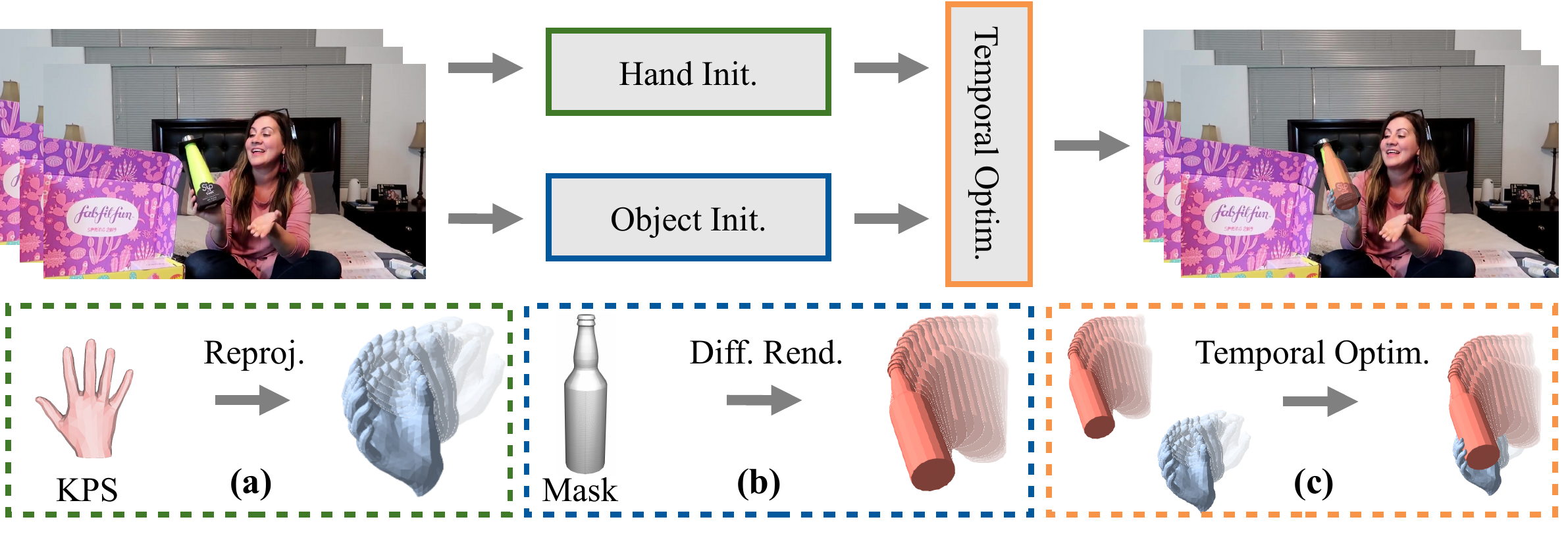}
\caption{\textbf{Reconstructing hands and objects from videos.} We present an optimization-based technique for reconstructing hands and objects from videos, leveraging spatial image cues (keypoints, masks, depth) and temporal smoothness constraints (4D, optical flow).}
\label{fig:vision}\vspace{0mm}
\end{figure*}
%##################################################################################################

\newpage

%%%%%%%%%%%%%%%%%%%%%%%%%%%%%%%%%%%%%%%%%%%%%%%%%%%%%%%%%%%%%%%%%%%%%%%%%%%%%%%%%%%%%%%%%%%%%%%%%%%
\section{Recovering Hands and Objects from Videos}\label{sec:rhov}
In this section, we present the Reconstructing Hands and Objects from Videos (RHOV) approach. RHOV takes a 3D object model and a monocular RGB video sequence as input and produces right hand and manipulated object poses in 3D for all frames of the video sequence. The hand poses are represented by articulation parameters in the MANO \cite{mano} format and a global translation. Object poses are represented by global rotation, translation and scale parameters (subject to scale ambiguity). This RHOV pipeline begins with independent hand and object reconstruction methods and then jointly optimizes hand and object poses across all frames of the video sequence.

\subsection{Hand Pose Estimation}\label{sec:hand_initialization}
Initial hand poses are produced with Frank MoCap \cite{frankmocap} which generates hand articulation parameters in the MANO \cite{mano} format from RGB frames. To generate a right hand track across all video frames, we start at the center of the sequence and generate a hand bounding box track both in the forward and backward directions through the frames. This is done by using the current hand box to initialize the box in the next frame. These boxes inform Frank MoCap where to produce a hand reconstruction. This method helps resolve ambiguities between multiple right hand instances and between the right and left hand detections.

\subsection{Object Pose Estimation}\label{sec:init_obj_pose}
Given a 3D object model, 3D object pose reconstructions are generated for every frame of the video sequence. This is accomplished by first producing 2D segmentation masks for the object of interest across all frames of the video sequence, referred to ask the object track. Next, the 3D object poses are generated from this 2D object track using gradient based optimization with a differentiable renderer.

\ourparagraph{Object mask tracking.} For each frame of the video, we use the hand-object detector from~\cite{100doh} to produce a bounding box for the object being manipulated by the right hand. Next, 2D masks are generated for all objects using the PointRend~\cite{pointrend} instance segmentation model for every frame. Object segmentation masks for the object of interest are determined by computing the intersection-over-union (IoU) between bounding boxes from each Detectron instance and the object box from the hand-object detector. The set of extracted masks is the object mask track which provides 2D cues about the 3D object pose.

%##################################################################################################
\begin{figure*}[t]\centering\vspace{0mm}
\includegraphics[width=1\linewidth]{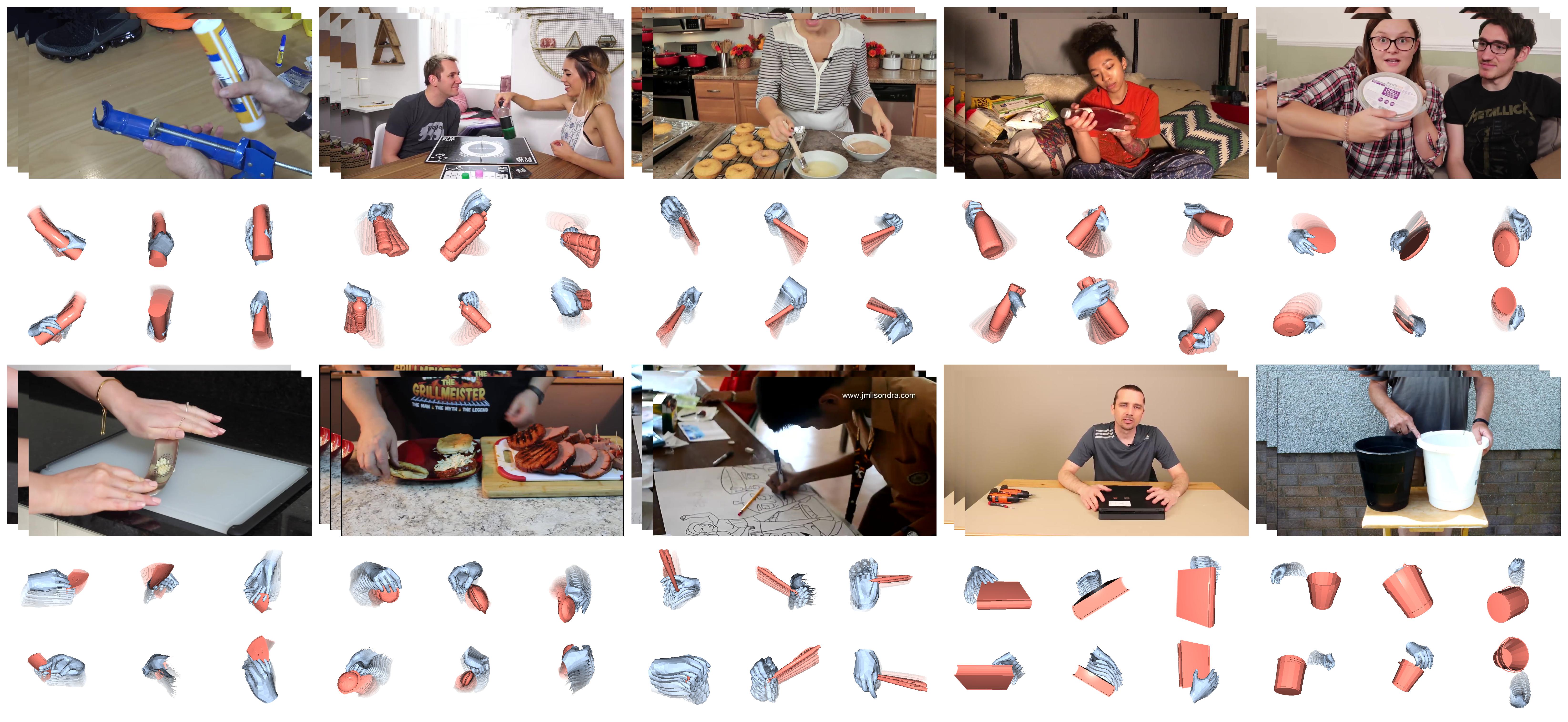}
\caption{\textbf{RHOV, qualitative results.} We show example reconstructions from RHOV. Each 4D reconstruction is shown from six views. }
\label{fig:vision_two}\vspace{-2mm}
\end{figure*}
%##################################################################################################

\ourparagraph{3D pose estimation.} To construct 3D object poses from the object mask track, we first sample random rotations $R$ and translations $T$ and score the transformed object pose using a series of loss functions. Given $\mathcal{R}$ is the rendering function, $s$ is the object scale, and $V$ are the object model vertices, we formulate a mask loss $\mathcal{L}_{proj}$ between the rendered object and the object mask $\mathcal{M}_{track}$. This loss is only computed for regions in which the hand mask generated from the initial hand pose (section \ref{sec:hand_initialization}) is not present to account for pose ambiguities due to the hand occluding the object.
\begin{equation}
    \mathcal{L}_{proj} = ||\mathcal{R}((RV+T)*s) - \mathcal{M}_{track}||_2^2
\end{equation}

We further leverage the MiDaS \cite{midas} approach to generate depth predictions $\mathcal{D}_{pred}$ from monocular RGB frames. Due to scale ambiguities from monocular RGB only sequences, we normalize depth to 0 mean and unit-variance within the bounds of the object with the normalization operator $\mathcal{N}$. Given $\mathcal{D}_{render}$ is the depth map generated by rendering the object, we evaluate depth loss as follows:

\begin{equation} \label{eq:depth_loss}
    \mathcal{L}_{depth} = ||\mathcal{N}(\mathcal{D}_{render}) - \mathcal{N}(\mathcal{D}_{pred}))||_2^2
\end{equation}

With these loss functions we formulate the following optimization problem with loss weights $\lambda_{proj}$ and $\lambda_{depth}$:
\begin{equation}\label{eq:init_obj_optim}
\begin{aligned}
\min_{R, T, s} \quad & \lambda_{proj}\mathcal{L}_{proj} + \lambda_{depth}\mathcal{L}_{depth}
\end{aligned}
\end{equation}

To minimize this objective we use a differentiable renderer \cite{neuralrenderer} and take a fixed number of gradient-steps with the Adam optimizer \cite{kingma2014adam}. Solving this problem yields an object pose for a single frame. We repeat this procedure for all frames by initializing the object pose to be the optimization result from the prior frame. This encourages smoothness for the object pose between frames and reduces issues with symmetrical ambiguities common in everyday objects.

\subsection{Temporal Optimization}
Given initial hand and object poses from \ref{sec:hand_initialization} and \ref{sec:init_obj_pose}, we formulate a joint optimization procedure involving the hand and object across all frames of the video sequence.

\ourparagraph{Optimization parameters.} The joint optimization procedure optimizes over the following variables: hand translation $T_h$ and rotation $R_h$, hand finger articulation $A_h$ in MANO \cite{mano} format, object rotation $R_o$ and translation $T_o$, and object scale $s_o$. All parameters have batch dimension $B$ except for $s_o$ which is shared among all frames. We use a MANO layer $\mathcal{M}\mathcal{L}$ as a differentiable implementation of the MANO model \cite{hasson19_obman} to map $A_h$ and $R_h$ to hand vertices. We compute transformed hand vertices $V_h$ and object vertices $V_o$ as follows: $V_h = \mathcal{M}\mathcal{L}(A_h) + T_h$ and $V_o = (R_oV_o + T_o) * s_o$. Our joint optimization procedure uses the following loss functions:

\textbf{Mask Loss.}
We evaluate 2D hand-object alignment with mask losses by comparing the mask from the renderer and the segmentation masks for the hand and object.

\textbf{Interaction losses.} 
To limit mutual hand-object mesh penetrations we use a signed distance field loss $\mathcal{L}_{p}$ between the hand and object meshes. To encourage interaction, we penalize the distance between the vertices on the finger tips and their nearest vertex on the object mesh as $\mathcal{L}_{inter}$.

\textbf{Depth loss.}
We use the same depth loss $\mathcal{L}_{depth}$ as in equation \ref{eq:depth_loss} to penalize object depth relative to the depth estimate from the RGB frame generated by \cite{midas}.

\textbf{Temporal loss.}
We encourage temporal smoothness across frames by penalizing parameter value changes between consecutive frames. This loss is applied to hand and object vertices, hand and object translation, and MANO parameters. For a given parameter at a certain frame index ${p_i}$, the loss is given by:

\begin{equation}
\begin{aligned}
\mathcal{L}_{temporal} = \sum\limits_{t=i-4}^{i} 0.5^{i-t}||p_i - p_t||_2^2
\end{aligned}
\end{equation}

\textbf{Optical Flow.}
We use the RAFT \cite{raft} model to generate optical flow predictions between consecutive frames as a reference for object motion. We denote $I_{of}^t$ as the optical flow image where $I_{of}^t(x)$ indicates the flow at pixel coordinate $x=(u, v)$ from frame $t-1$ to frame $t$. To compare to this reference flow, we project object vertices $V_o$ onto the image plane using the camera projection operator $\Pi$. These projections are then filtered according to visibility operator $\mathcal{V}(v)$ which is an indicator function for whether object vertex $v$ is visible in the rendered image. This is done as non-visible vertices shouldn't be compared to the optical flow reference which only accounts for visible parts of the object. Given that $V_o$ is the set of object vertices, We compute the loss $\mathcal{L}_{of}$ for frame $t$ as:

\begin{equation}
\begin{aligned}
\mathcal{L}_{of} = \sum\limits_{v \in V_o} \mathcal{V}(v)||\Pi(v^t) - \Pi(v^{t-1}) - I_{of}^t(\Pi(v^t))||_2^2
\end{aligned}
\end{equation}

\newpage

\textbf{Occlusion Loss.}
The 3D hand-object reconstruction should obey visibility constraints induced by mutual hand-object occlusion. We denote the visible 2D mask of the object by $\mathcal{M}_o^{vis}$, which can be directly obtained from the object track. Given the hand mask $\mathcal{M}_h$ generated from rendering the initial hand pose from section \ref{sec:hand_initialization}, the visible hand mask $\mathcal{M}_h^{vis} = \mathcal{M}_h - \mathcal{M}_h \cap \mathcal{M}_o$. This is due to the observation that segmentation masks for the object only can include visible parts of the object and that the hand mask may overlap with sections of that object mask. We use the renderer to generate depth maps $\mathcal{D}_h$ and $\mathcal{D}_o$ that can be used to identify hand-object visibility for every image coordinate for the current hand-object pose. We compute an occlusion score for hand $\mathcal{L}_{occ}^h$ and object $\mathcal{L}_{occ}^o$ that penalizes regions in which the hand is in front of the object when it shouldn't have been and vice-versa.

\textbf{MANO Loss.}
To penalize unnatural MANO~\cite{mano} hand articulation parameters $A_h$, we penalize the norm of the latent space parameter vector: $\mathcal{L}_{mpn}=||A_h||_2^2$.

\textbf{GrabNet Loss.}
The GrabNet \cite{grabnet} approach takes in hand pose estimates in MANO format and object pose estimates and generates MANO parameters $A_{grabnet}$ for a realistic grasp of the specified object. We penalize the norm between GrabNet parameters and the current articulation parameters to encourage similarity to the GrabNet parameters: $\mathcal{L}_{gn} = ||A_{grabnet} - A_h||_2^2$. This loss is only applied to frames where the hand is within a distance threshold of the object.

\textbf{Projection Loss.}
To encourage stability in the optimization procedure, we penalize deviations for image plane projection between the initial and current object poses with $\mathcal{L}_{iop}$ ($\mathcal{L}_{ihp}$ exists for hand). Given $v_o^i$ indicates the initial 3D object vertex location, we compute:
\begin{equation}
\begin{aligned}
\mathcal{L}_{iop} = \sum\limits_{v_o \in V_o} ||\Pi(v_o) - \Pi(v_o^i)||_2^2
\end{aligned}
\end{equation}

\ourparagraph{Optimization procedure.} To perform the joint optimization, we formulate an optimization problem similar to equation \ref{eq:init_obj_optim} except with parameters and loss functions defined in this subsection. We perform this optimization in two stages. In the first stage, we attempt to bring the object close to the hand as their initialization as described in sections \ref{sec:hand_initialization} and \ref{sec:init_obj_pose} may be far apart from one another. To accomplish this we only optimize over $T_o$ and $s_o$ subject to only $\mathcal{L}_{iop}$ and $\mathcal{L}_{inter}$. The idea is to keep the hand fixed, and adjust the object to be physically close ($\mathcal{L}_{inter}$), while retaining the sample image plane projection ($\mathcal{L}_{iop}$).

The second phase of this procedure optimizes over all parameters subject to all of the loss functions. Halfway through the second stage of optimization, GrabNet \cite{grabnet} is run as the hand is typically in a reasonable pose with respect to the object, which yields better GrabNet results. The Adam optimizer \cite{kingma2014adam} is used for both optimization stages.

\newpage

%%%%%%%%%%%%%%%%%%%%%%%%%%%%%%%%%%%%%%%%%%%%%%%%%%%%%%%%%%%%%%%%%%%%%%%%%%%%%%%%%%%%%%%%%%%%%%%%%%%
\section{Imitating Object Interactions}\label{sec:rl}

%##################################################################################################
\begin{figure}[t]\centering\vspace{0mm}
\includegraphics[width=1\linewidth]{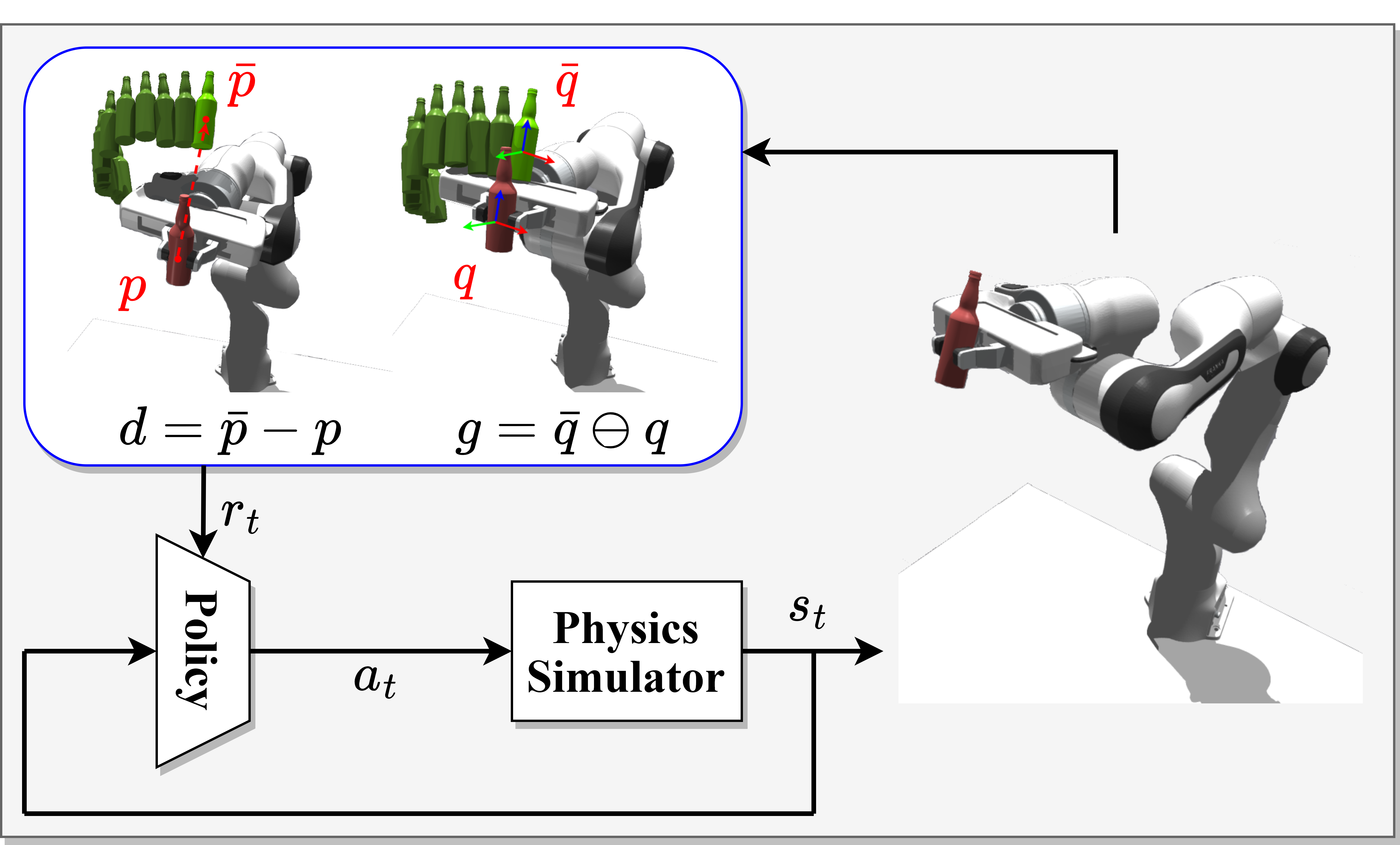}
\caption{\textbf{Imitating object interactions.} We frame the task of imitating the observed object interaction as an RL problem. The recovered 4D object trajectory enables us to compute dense rewards, based on the object pose distance, and train a policy to imitate it.}
\label{fig:rl}\vspace{-2mm}
\end{figure}
%##################################################################################################

In this section we describe our procedure for learning to imitate object interactions in a physics simulator.

%##################################################################################################
\begin{figure*}[t]\centering\vspace{0mm}
\includegraphics[width=1\linewidth]{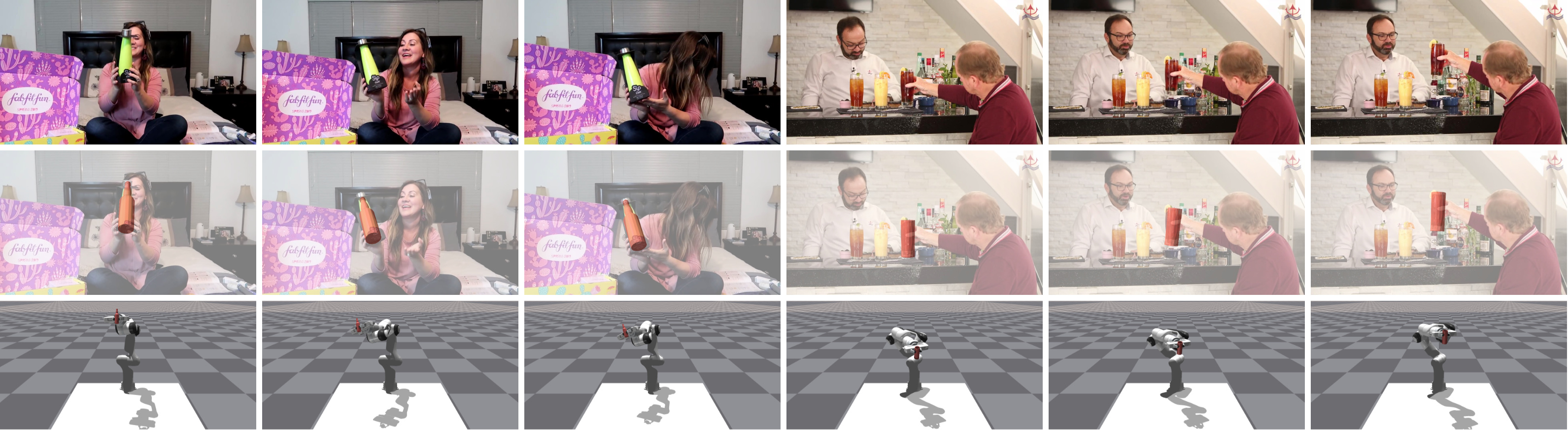}
\caption{\textbf{Reconstruction and imitation results per step.} For two input video sequences (top), we show the projections of RHOV object reconstructions onto the image (middle), and the corresponding frame being imitated by a robot in a physics simulator (bottom).}
\label{fig:rl_qual}\vspace{-2mm}
\end{figure*}
%##################################################################################################

\subsection{Task Setup}

In order to imitate an object interaction, we require a simulation environment with the appropriate workspace and objects. We place a Franka robotic arm on a table and import the object model. For each object, we manually estimate the physical object properties (\eg, mass).

The recovered trajectories start and end at arbitrary points of the video clip. In most cases, at the start of the trajectory, the object is in the hand, being manipulated or carried around, rather than being stationary on a support surface (\eg, table). In order to imitate the interaction, we must first bring the object to an appropriate starting position. We thus divide the task into two phases. In phase one, the goal is to pick up the object from a starting position on the table and bring it to the initial pose of the recovered trajectory. In phase two, the goal is to imitate the remainder of the object trajectory, matching the object pose at each step.

\subsection{Learning to Imitate}

We adopt an object-centric perspective and focus on imitating the object trajectory without the hand. This makes our approach morphology-agnostic and enables us to use different embodiments, like a robotic arm with a parallel jaw gripper. We formulate the task as a reinforcement learning problem and describe main design choices next.

\ourparagraph{Goal specification.} The recovered 4D trajectory serves as a natural goal specification. Each trajectory is a sequence of position and rotation pairs. Reaching the goal requires matching the 6 DoF object pose at each of the timesteps. We consider a timestep to be matched when both the position and rotation are within a small distance from the target.

\ourparagraph{Actions.} Our control policy is a neural network with 3 hidden layers. We make no assumptions about the robot or the problem domain and perform control in joint position space. Specifically, we predict delta joint positions at 60 Hz.

\ourparagraph{Observations.} We provide the robot joint positions and velocities, distance from the gripper to the object, and the position and rotation distance to the current goal pose. We also include the goal pose of the current and next timestep in the target object trajectory. When the current goal pose is reached, we advance to the next goal pose in the sequence.

\ourparagraph{Reward function.} Our reward function consists of two phases, intial ($\sigma=0$) and imitation ($\sigma=1$):
\begin{equation}
\begin{aligned}
r_t = (1 - \sigma) \cdot w_1 \cdot r_{\text{initial}} + \sigma \cdot w_2 \cdot r_{\text{imitation}}
\end{aligned}
\end{equation}
The initial reward aims to lift the object off of the table and bring it to first target frame, and is as follows:
\begin{equation}
\begin{aligned}
r_{\text{initial}} = w_{\text{initial}}^T [r^{lf}_t, r^{rf}_t, h_t, h_t \cdot r^x_t, h_t \cdot r^y_t, b^x, b^y, -a]
\end{aligned}
\end{equation}
The first two terms $r^{lf}_t$ and $r^{rf}_t$ incentivize the fingers of the parallel jaw gripper to move close to the object.
\begin{equation}
\begin{aligned}
r^{lf}_t, r^{rf}_t = \frac{1}{0.04 + \big \| \overline{p_{lf, rf}} - p_{\text{obj}} \big \|_2}
\end{aligned}
\end{equation}
Next, $h_t$ takes on a value 1 if the object is above the table and 0 otherwise. When the object is above the table, reward terms $r^x_t$ and $r^y_t$ are added. The term $r^x_t$ encourages the current position of the object to match that of the first target frame. A bonus of $b^x$ is received if the difference in positions $x = \bar{p} - p$ falls below a threshold. Similarly, $y_t$ encourages the rotational difference between the object and first target frame to match, with an associated bonus $b^y$ when the angular magnitude of the quaternion difference $y = \bar{q} \ominus q$ is below the threshold. 
\begin{equation}
\begin{aligned}
r^x = \big \| \bar{p} - p \big \|_2, \;\; r^y =  \bar{q} \ominus q   
\end{aligned}
\end{equation}

A small action penalty of $-a$ is added at the end of the initial reward.
Once the initial frame of the trajectory is reached, the imitation reward encourages the robot to imitate the rest of the trajectory, and is given by:
\begin{equation}
\begin{aligned}
r_{\text{imitation}} = w_{\text{im}}^T [r^d_t, r^{\dot{d}}_t, r^g_t, r^{\dot{g}}_t, b^d, b^{\dot{d}}, b^g, b^{\dot{g}}, f_t, b^f]
\end{aligned}
\end{equation}
The position reward $r^d_t$ encourages position of the object at the current timestep to match the object's position in the current target frame. A bonus $b^d$ is added when the magnitude of the difference between 3D positions $d = \bar{p} - p$ falls under a certain threshold.
\begin{equation}
\begin{aligned}
r^d_t = \text{exp}\Big(-k_d \sum_{i} \big \| p_{\text{target}} - p_{\text{curr}} \big \|_2^2 \Big)
\end{aligned}
\end{equation}
Similarly, the orientation reward $r^g_t$ encourages orientation of the object at the current timestep to match the object's orientation in the current target frame. A bonus $b^g$ is added when the angular magnitude of the quaternion difference $g = \bar{q} \ominus q$ reaches a threshold.
\begin{equation}
\begin{aligned}
r^g_t = \text{exp}\Big(-k_g \sum_{i} \big \| q_{\text{target}} \ominus q_{\text{curr}} \big \|_2^2 \Big)
\end{aligned}
\end{equation}
We also include terms $r^{\dot{d}}_t$ and $r^{\dot{g}}_t$ that incentivize the linear and angular velocity of the object to match those of the current target frame. A corresponding bonus $b^{\dot{d}}$ or $b^{\dot{g}}$ is added when the difference in velocity $\bar{\dot{d}}-\dot{d}$ or $\bar{\dot{g}}-\dot{g}$ falls under the threshold. The target linear and angular velocities are computed using finite differences.
\begin{equation}
\begin{aligned}
r^{\dot{d}, \dot{g}}_t = \text{exp}\Big(-k_{\dot{d}, \dot{g}} \sum_{i} \big \| v_{\text{target}} - v_{\text{curr}} \big \|_2^2 \Big)
\end{aligned}
\end{equation}
Finally, we include a term $f_t$ that measures the percentage of reached frames and increases the rewards as a function of the frames in the target trajectory have been matched, with an associated bonus $b^f$ every time a target frame is reached.

\ourparagraph{Training.} We construct our tasks in the PixMC~\cite{Xiao2022} benchmark suite, powered by the NVIDIA IsaacGym simulator~\cite{makoviychuk2021isaac}.  We train a policy per video using PPO~\cite{schulman2017proximal}.

\clearpage

%%%%%%%%%%%%%%%%%%%%%%%%%%%%%%%%%%%%%%%%%%%%%%%%%%%%%%%%%%%%%%%%%%%%%%%%%%%%%%%%%%%%%%%%%%%%%%%%%%%
\section{Experiments}\label{sec:experiments}

%##################################################################################################
\begin{figure}[t]\centering\vspace{0mm}
\includegraphics[width=0.95\linewidth]{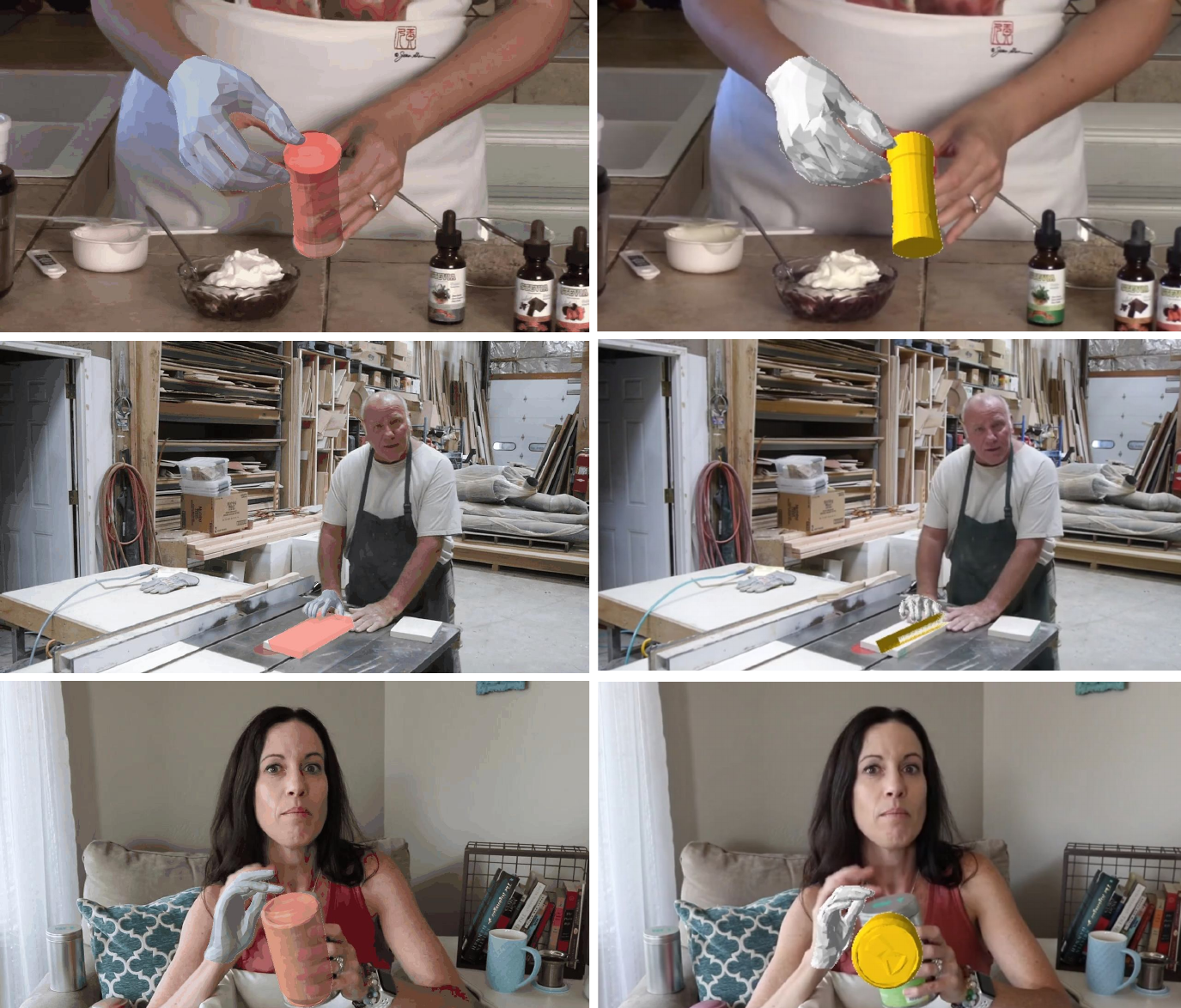}
\caption{\textbf{Qualitative, Internet videos.} We compare our RHOV method (left) to the state-of-the-art approach HOman~\cite{homan} (right).}\vspace{-1mm}
\label{fig:qual_compare}\vspace{-2mm}
\end{figure}
%##################################################################################################

We now evaluate our RHOV approach. We show qualitative results on Internet videos, quantitative results on in-the-lab videos with ground truth labels, and perform ablation studies of different components of our approach.

\subsection{Qualitative results}

\ourparagraph{Video data.} We use the MOW~\cite{rhoi} dataset as the source of Internet videos. MOW consists of 500 images that are annotated with 3D hands and objects. The images in MOW are sourced from the 100 Days of Hands Videos~\cite{100doh} dataset which contains YouTube videos of people interacting with objects. For each MOW image, we extract a 60 frame chunk of the video centered at the annotated frame. For the single frame annotated in MOW, we initialize our object track with the provided bounding box around the object of interest. Rather than randomly sampling poses for the first frame, we use the object pose and model provided by MOW to initialize our pose estimate.

\ourparagraph{Reconstructions.} In Figure~\ref{fig:vision_two}, we show example reconstructions obtained with the RHOV technique. For each input video, we show six different views of the 4D hand-object trajectory. We see that our approach is able to handle various interactions and objects. Please also check our~\href{https://austinapatel.github.io/imitate-video}{project page} for video results. In total, we "4Dfy" 100 video sequences and will release all of the trajectories.

\ourparagraph{Comparisons.} In Figure~\ref{fig:qual_compare}, we compare RHOV reconstructions to a state-of-the-art technique for reconstructing hands and objects from videos. We observe that RHOV produces reconstructions of better quality on Internet videos.

%\newpage

\subsection{Quantitative results}

We further evaluate our technique on in-the-lab datasets where ground truth annotations are available (measured with multiple calibrated depth cameras).

%##################################################################################################
\begin{table}[t]\centering
\resizebox{0.65\columnwidth}{!}{\tablestyle{6pt}{1.1}
\begin{tabular}{@{}l|cc@{}}
    & Hand Error $^{\downarrow}$ & Object Error $^{\downarrow}$ \\
\shline
% \hline
 Photo-consist~\cite{hasson20_handobjectconsist} & 14.7 & 26.6 \\
 RHO~\cite{rhoi} & 9.7 & 19.9 \\
 HOman~\cite{homan}  & 17.3 & 5.4 \\
 RHOV (ours) &  10.1 & 9.6 \\
\end{tabular}}\vspace{-1mm}
\caption{\textbf{RHOV quantitative comparisons, HO3D.} Overall, our approach is comparable to the state-of-the-art techniques.}
\label{tab:eval_ho3d}\vspace{-1mm}
\end{table}
%##################################################################################################

\ourparagraph{HO3D.} We first evaluate our approach on HO3D~\cite{ho3d_honnotate}, which contains 3D annotations for hands and objects for 68 sequences across 10 objects. We follow the experimental protocol from~\cite{hasson20_handobjectconsist} which evaluated on the MC2 sequence. We use the evaluation metrics from \cite{rhoi}, which are mean per joint position error (MPJPE) for hand (after global pose alignment) and chamfer distance for object (after global translation alignment).

We report the results in Table~\ref{tab:eval_ho3d}. We observe that our approach considerably outperforms the per-frame method in~\cite{rhoi} which highlights the benefits of using temporal information. We further compare our approach to two state-of-the art approaches that leverage video as well. We see that our approach performs favorably while producing better reconstructions on Internet videos (Figure~\ref{fig:qual_compare}).

%##################################################################################################
\begin{table}[t]\centering
\resizebox{0.7\columnwidth}{!}{\tablestyle{6pt}{1.1}
\begin{tabular}{@{}l|cc@{}}
    & Root-Relative $^{\downarrow}$& Procrustes$^{\downarrow}$ \\
\shline
 A2J \cite{dexycb_citation45_a2j} & 23.9 & 12.1 \\
 \cite{dexycb_citation31_weaklysupervised} + HRNet32 & 17.3 & 6.8 \\
  RHOV (ours) & 32.8 & 10.4 \\
\end{tabular}}\vspace{-1mm}
\caption{\textbf{RHOV quantitative comparisons, DexYCB hand.} Numbers reported are in MPJPE error (mm) for the S0 split.}
\label{tab:eval_dexycb_hand}\vspace{-1mm}
\end{table}

\begin{table}[t]\centering
\resizebox{0.6\columnwidth}{!}{\tablestyle{6pt}{1.1}
\begin{tabular}{l|c}
    & Average Recall (\%) $^{\uparrow}$ \\
\shline
 PoseCNN~\cite{Xiang2018}  & 41.7\\
 CosyPose~\cite{dexycb_citation19_cosypose} & 60.0\\
 RHOV (ours) & 41.2 \\
\end{tabular}}\vspace{-1mm}
\caption{\textbf{RHOV quantitative comparisons, DexYCB object.} Numbers reported are in average recall for the S0 split of \cite{dexycb}. We achieve respectable performance despite not training on DexYCB.}
\label{tab:eval_dexycb_object}\vspace{-3mm}
\end{table}

%##################################################################################################

\ourparagraph{DexYCB.} Next, we report results on DexYCB~\cite{dexycb}, which is an in-the-lab dataset with 1000 sequences captured from eight depth cameras. Each video sequence shows a human hand picking an object from the table. In Table~\ref{tab:eval_dexycb_hand} and Table~\ref{tab:eval_dexycb_object}, we report results for DexYCB hand and object evaluation, respectively. We report results on the S0 split following the procedure from the paper after aligning with ground truth translation (S1 results in supplement). We compare to the well-tuned and state-of-the-art methods from the DexYCB paper. Our approach achieves respectable performance despite not being trained on the DexYCB dataset.

\clearpage

\subsection{Ablation studies}

%##################################################################################################
\begin{table}[t]%\vspace{-3mm}
\centering
\subfloat[\textbf{Temporal optimization improves object smoothness.} Temporal distance is computed as average vertex L2 distance between adjacent frames. \label{tab:ablation_temporal}]{
\makebox[1\linewidth]{
\tablestyle{10pt}{1.1}
\begin{tabular}{@{}r|cc@{}}
    & Object Error $^{\downarrow}$ & Temp. Dist. $^{\downarrow}$ \\
\shline
  per frame & 14.3 & 87.3 \\
  temporal & 13.0 & 4.1 \\
\end{tabular}
}}\\[2mm]
\subfloat[\textbf{Better 2D masks lead to better 3D reconstructions.} We observe that ground truth 2D object masks lead to better 3D object reconstructions. This suggests that our 3D approach may improve with better 2D models.\label{tab:ablation_mask}]{
\makebox[1\linewidth]{
\tablestyle{10pt}{1.1}
\begin{tabular}{@{}l|cc@{}}
    & Object Error $^{\downarrow}$ \\
\shline
 predicted masks & 13.0 \\
 oracle masks  & 9.3 \\
\end{tabular}
}}\\[2mm]
\subfloat[\textbf{Reconstructing the hand improves the object.} We see that jointly optimizing
both the hand and the object leads to better object reconstructions. \label{tab:ablation_joint_ho}]{
\makebox[1\linewidth]{
\tablestyle{10pt}{1.1}
\begin{tabular}{@{}l|cc@{}}
    & Object Error $^{\downarrow}$\\
\shline
 independent optim. & 14.6 \\
 joint optim. & 13.0 \\
\end{tabular}
}}\vspace{-4mm}
\caption{\textbf{Ablation experiments on reconstruction technique.} We perform ablation experiments for different aspects of the RHOV method. All ablation numbers are averaged over the 13 evaluation sequences from the HO3D dataset~\cite{ho3d_honnotate}.}
\label{tab:rhov_ablations}\vspace{-2mm}
\end{table}
%##################################################################################################

Finally, we perform ablation experiments to understand the impact of different components of our approach.

\ourparagraph{Temporal smoothness.}  In Table~\ref{tab:ablation_temporal}, we study the impact of temporal optimization on the quality of RHOV reconstructions. We compare our default settings to a variant that independently reconstructs hand and object per frame. We observe that temporal optimization greatly smooths object movement. Qualitatively, the object often flips orientation between frames in the per frame approach due to symmetrical ambiguities in the object model, while in temporal optimization it does not. Thus temporally smooth reconstruction methods like RHOV are needed to produce valid trajectories for imitating object pose with a robot.

\ourparagraph{2D mask quality.} RHOV object reconstruction relies on 2D object masks computed using a pre-trained instance segmentation PointRend~\cite{pointrend} model. In particular, we estimate the object pose via differentiable rendering with predicted 2D masks. In Table~\ref{tab:ablation_mask}, we evaluate our method using oracle masks available as part of HO3D. We observe that oracle 2D masks improve the object reconstruction performance. This may suggest that the 3D reconstruction from Internet videos may improve with better 2D segmentation models.

\ourparagraph{Impact of object on hand.} Our approach imitates the reconstructed object trajectories without taking hand information into account. This may suggest that there is no need to reconstruct the hands. However, we observe that reconstructing the hand improves object reconstructions. We report results in Table~\ref{tab:ablation_joint_ho}. This suggests that joint hand-object reconstruction is useful even if we only need the object.

\newpage

%##################################################################################################
\begin{table}[t]\centering
\resizebox{\columnwidth}{!}{\tablestyle{10pt}{1.1}
\begin{tabular}{@{}l|cc|c@{}}
    & Pos. Dist.$^{\downarrow}$  & Rot. Dist.$^{\downarrow}$ & Success (\%)\\\shline
 complete reward    & 0.039 & 0.289    & 79.3 \\
 w/o velocity & 0.098 & 0.683 & 46.6 \\
 w/o num. matched   & 0.099 & 0.725 & 30.9 \\
 w/o rotation  & 0.070 & 1.782 & 0.00 \\
\end{tabular}}\vspace{-1mm}
\caption{\textbf{Ablation experiments on imitation rewards.} We observe that removing different imitation reward components leads to considerable drops in overall performance.}
\label{tab:ablation_reward}\vspace{-4mm}
\end{table}
%##################################################################################################

\begin{figure}[h]\centering\vspace{3mm}
\includegraphics[width=0.95\linewidth]{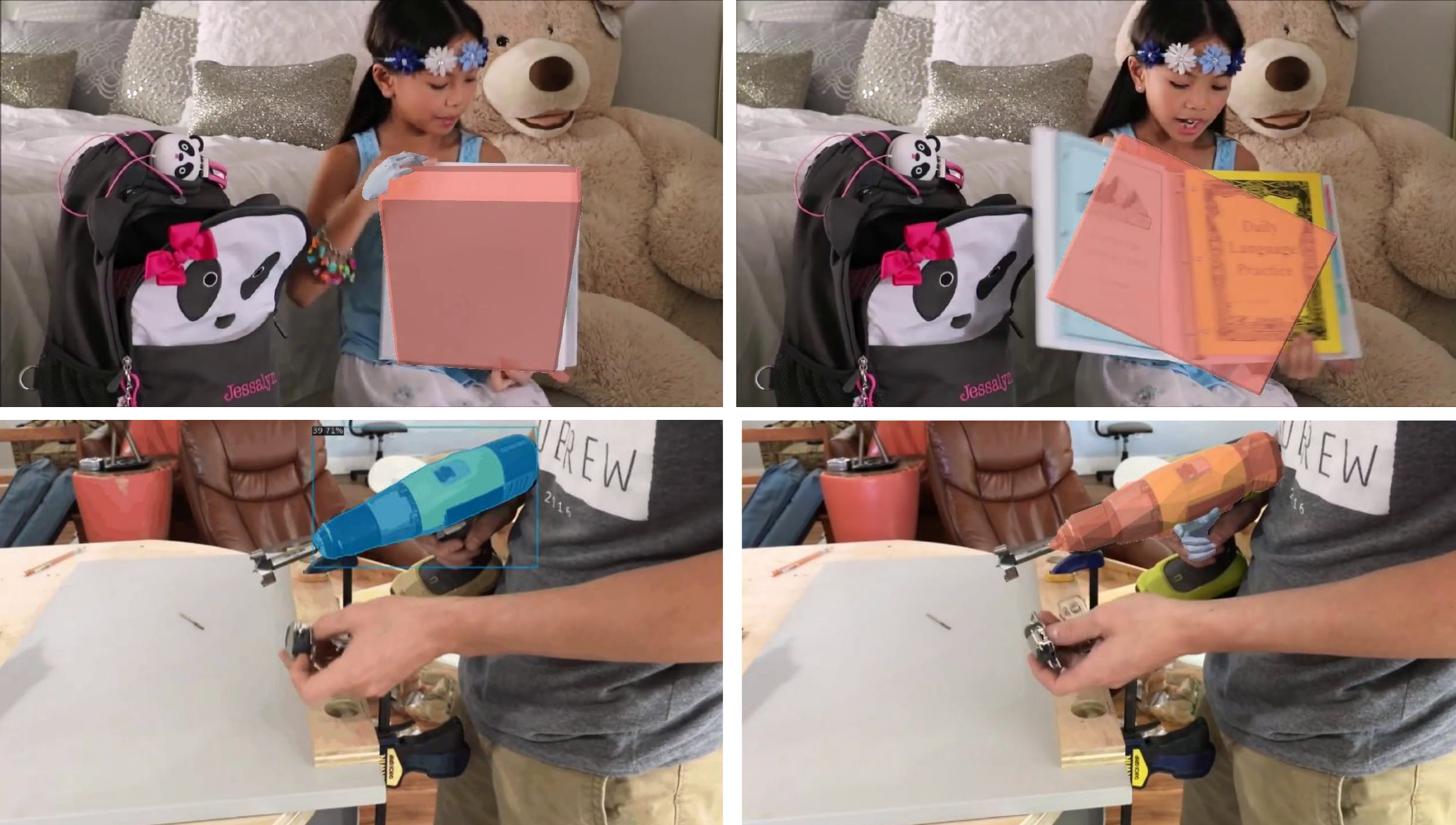}
\caption{\textbf{Example failure cases of RHOV.} We show failures due to the object model mismatch (top) and poor 2D masks (bottom).}
\label{fig:qual_failure}\vspace{-2mm}
\end{figure}
%##################################################################################################

\ourparagraph{Reward terms.} Next, we study the imitation aspect of our approach. Our approach uses reinforcement learning to match the recovered 4D trajectory. We use the recovered trajectory to compute a dense reward function. The choice of the reward function plays a key role in the success of the approach. Our reward consists of several terms (\eg, rotation distance). In Table~\ref{tab:ablation_reward}, we report performance with one of the reward terms removed. We observe that removing any of the ablated terms leads to a drop in performance suggesting that all terms play an important role in our approach.

\ourparagraph{Failure cases.} We show a few representative failure cases of the RHOV technique. The reconstruction fails in cases with poor 2D masks, incorrect object models, articulated object models, and severe hand-object occlusions.

%%%%%%%%%%%%%%%%%%%%%%%%%%%%%%%%%%%%%%%%%%%%%%%%%%%%%%%%%%%%%%%%%%%%%%%%%%%%%%%%%%%%%%%%%%%%%%%%%%%
\section{Conclusion}

We study the problem of imitating object interactions from Internet videos. We make two main contributions: (1) a reconstruction technique RHOV (Reconstructing Hands and Objects from Videos), which reconstructs 4D hand-object trajectories by leveraging 2D image cues and temporal smoothness constraints; (2) a system for imitating object trajectories in a physics simulator with reinforcement learning. Please see our~\href{https://austinapatel.github.io/imitate-video}{project page} for video results.

\ourparagraph{Acknowledgments.} We thank Shubham Goel, Georgios Pavlakos, and Jathushan Rajasegaran for discussions. This research was supported by DARPA Machine Common Sense and ONR MURI (N00014-21-1-2801) programs.

\clearpage

%%%%%%%%%%%%%%%%%%%%%%%%%%%%%%%%%%%%%%%%%%%%%%%%%%%%%%%%%%%%%%%%%%%%%%%%%%%%%%%%%%%%%%%%%%%%%%%%%%%
{\small\bibliographystyle{ieee_fullname}\bibliography{references}}

\end{document}